\documentclass[11pt,twocolumn]{article}
\usepackage[parfill]{parskip}
\usepackage{listings}
\usepackage{color}
\usepackage{gensymb}
\usepackage{fancyhdr} % for headers and footers
\usepackage{graphicx}
\usepackage{framed}
\usepackage{braket}
\usepackage{multicol}
\usepackage{hyperref}
\usepackage{amsfonts}
\usepackage{mathtools}
\usepackage[top=1.0in, bottom=0.75in,left=0.8in,right=0.8in]{geometry}

\usepackage{amsmath}
\usepackage{amsthm}

\usepackage{mathrsfs}
\usepackage{abstract}

\definecolor{dkgreen}{rgb}{0,0.6,0}
\definecolor{gray}{rgb}{0.5,0.5,0.5}
\definecolor{mauve}{rgb}{0.58,0,0.82}

	\title{Learning the Correction for Multi-Path Deviations in Time-of-Flight Cameras}
\author{
        Mojmir Mutny\thanks{The University of Edinburgh;M.Mutny@sms.ed.ac.uk} \thanks{This work has been supported by DAAD RISE Scholarship} , Rahul Nair\footnote{HCI - Heidelberg University} , Jens-Malte Gottfried\footnotemark[2]  \\
}
\date{\today}

\begin{document}

\twocolumn[
\begin{@twocolumnfalse}

\maketitle

 \begin{abstract}
  \vspace{1cm}
 The Multipath effect in Time-of-Flight(ToF) cameras still remains to be a challenging problem that hinders 
 further processing of 3D data information. Based on the evidence from previous literature, we 
 explored the possibility of using machine learning techniques to correct this effect. Firstly, we 
 created two new datasets of of ToF images rendered via ToF simulator of LuxRender. These two datasets contain corners in multiple orientations and with different material properties. We chose scenes with corners as multipath effects are most pronounced in corners. Secondly, we used this dataset to construct a learning model to predict real valued corrections to the ToF data using Random Forests. We found out that in our smaller dataset we were able to predict real valued correction and improve the quality of depth images significantly by removing multipath bias. With our algorithm, we improved relative per-pixel error from average value of 19\% to 3\%. Additionally, variance of the error was lowered by an order of magnitude.
 \vspace{1cm}
 \end{abstract}
\end{@twocolumnfalse}
]
\saythanks
\section{Introduction}
Time of Flight (ToF) cameras are one of the few possible models for successful depth measurement of a scene that accompany the usual 2D technology. The usual ToF camera contains an illumination light which emits IR light of stable frequency $f_m$ (usually $30 kHz$) \cite{Lefloch2013}. These beams are reflected from objects that are being captured and returned back to the camera recording chip. Here, according to the phase shift at a given pixel on the chip the depth of a pixel can be calculated. ToF technology is becoming increasingly more popular and has recently penetrated to mainstream marked via Kinect device for Xbox manufactured by Microsoft \cite{Wired}.
 
Although very successful, Time of Flight cameras are prone to a lot potential errors. One of these is a multiple return problem \cite{Grzegorzek2013}. In essence, this means that light incident to a single pixel on a chip does not come from a single reflection point but from multiple points. Thus light beams incident from distinctive reflection points have different phase shifts making it difficult to interpolate the right depth. Through clever processing this effect can be mitigated as amplitudes of unwanted reflections are differ from the desired. Thus, this problem can be interpreted in this scenario as noise removal problem. A prime example where this effect is greatly pronounced is when capturing sharp corners made of specular materials. In this instance, light from one plane of the corner is reflected to the other and vice versa creating a difficult scene to measure correctly.

\section{Background}
Multiple approaches have been proposed to reduce the error of measurement due to multipath effect. Some of them focus on the hardware level but many on software post-processing as well. For example, multipath problem can be partially improved by increasing sampling rate, but to this day, this remains a significant problem in time of flight imagining, and many solutions try to tackle it \cite{Grzegorzek2013,Freedman2014}. One of the more recent attempts to mitigate this effect was to use random forests to learn confidence of the pixels in \cite{Reynolds2011} and  \cite{Song2014}. Additionally, a general technique based on variational methods has been recently proposed in \cite{Freedman2014}. 

This work explores a potential application of large scale machine learning to alleviate this unwanted effect. In multipath correction learning, one of the biggest problems is that ground-truth depth datasets are very scarce if existing at all. In addition, if they are present, they usually contain only a small set of images due to difficulty of their precise measurement such as LiDar variant in \cite{Reynolds2011}. Another approach present in literature is to use ToF device itself to measure ground truth depth and average depth values over time \cite{Song2014}. However, this approach is most likely unfruitful if the position of camera is not changing as the multipath effect is a systematic bias not a statistical error. Even if the camera is moving it is difficult to present a certificate that such method works reliably.  

Without the ground truth depth information, we are cannot train the learning model reliably to recognize the multipath effect. Reynolds et. al. \cite{Reynolds2011} obtained their dataset by measuring the ground-truth images with very accurate LiDar device, and then converted the image to appropriate form to resemble the ground truth image for the camera. This is a very expensive and time consuming approach to generate a precise dataset for learning. We wanted to operate our learning model on much larger dataset, therefore, we chose to generate an artificial set of ToF images. Previously, a simulator for ToF cameras has been created based on popular LuxRender program that produces reliable simulated ToF images with ground truth \cite{Meister2013}. The scenes for LuxRender were modelled in open-source program Blender.

\section{Generating the database}
As multipath effect exhibits itself in many circumstances we chose to generate scenes where the effect is most pronounced and of greatest challenge. The best candidate that was already mentioned is a scene that contains corners. We chose to create a database with various corners (different angle and materials) in multiple positions and use this dataset for algorithm to learn the correction to the multipath error. The database\footnote{The database can be provided upon request. It is has not been made public due to its size.} is split to two different subsets \textit{challenging} and \textit{simple}. Most of the currently presented work deals with \textit{simple} database only. In future, we plan to work with the \textit{challenging} dataset as well.

Material model used to describe the surface of the corners are important factor that in influencing the validity of artificial ToF scenes. Some of the models provide better and more realistic results as shown in \cite{Kondermann2008}. However, as a general rule, the more complicated model is, the longer it takes to render it. Thus, one has to find balance between performance and accuracy. We chose to use fairly simple Ward anisotropic model \cite{Geisler-Moroder2010,Ward1992}, which has 4 parameters. Ward model assumes that the resulting material is weighted sum of perfectly lambertian surface and specular surface. Specular surface is characterised by it roughness. This means that reflective angles are distributed normally around the incidence angle with parameter $\sigma$ - specular roughness. Each of the surfaces (lambertian and specular) has its colour ($K_d$ and $K_s$ respectively), where the specular colour is set to $K_s=1.0$ all the time. Lastly a mixing parameter $\mu$ mixes the two models. All parameters $\sigma, \mu, K_d, K_s$ belong to $[0,1]$.

A database of parameters for different materials in Ward Model can be found in BRDF Material database maintained by CAVE laboratory at Columbia \cite{CAVEdat}. We picked a six materials from the database to give an idea of how are the parameters varying for real surfaces. They are presented in Figure \ref{fig:Materials_table}.
 
\begin{figure}
\centering
\begin{tabular}{ | l | l | l | l | l |}
    \hline
    Material Name & $\sigma$ & $\mu$ & $K_s$ & $K_d$ \\ \hline
     
	Concrete & 0.600672& 0.668533& 1.0& 0.994044 \\ \hline
	Wood& 0.598438& 0.132031& 1.0 & 0.965061\\ \hline
	Rough Plastic& 0.278057& 0.480943 & 1.0 & 0.969021 \\ \hline
	Limestone& 0.413544 & 0.292841 & 1.0 & 0.972684 \\ \hline
	Rough Paper& 0.311376 & 0.644926 & 1.0  & 0.937665 \\ \hline
	Foil & 0.252702 & 0.581514 & 1.0 & 0.891302 \\ \hline

\end{tabular}
\caption{Six Materials that were used to generate the simple data set. Data was taken from Columbia dataset \cite{CAVEdat}.} 
\label{fig:Materials_table}
\end{figure}

The \textit{challenging} database contains $10,000$ corners where 2-planes intersect. Each plane has its own material. Materials parameters, angles of the planes, and camera position have been uniformly sampled over the whole domains at random. This dataset contains very diverse and uncommon corners and camera angles. The camera always looks in the centre of the two plane intersection and is separated by 3 length (with cca. 7.5 limit of ToF device) units.

In addition to $10,000$ 2-plane corners, \textit{challenging} database contains also $10,000$ 3-plane corners where one plane is always perpendicular to the other two. Again each plane has its own material and parameters were sampled uniformly over their full domain. Specular colour is $1.0$ in both cases.

\begin{figure}
    \centering
    \includegraphics[width=7.5cm,height=4.5cm]{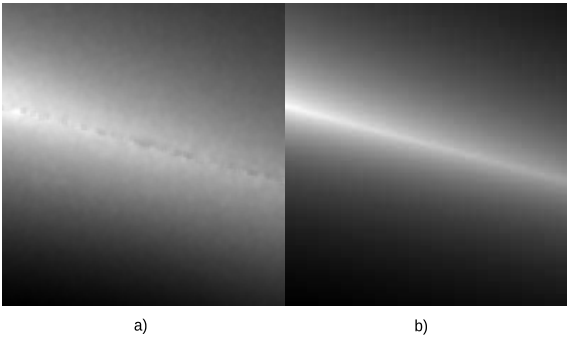}
    \caption{Depth Image of a corner from the \emph{simple} database. In a) one can see measured depth and in b) the ground truth image.The shade of grey determines the depth with maximum being 3.5 length units (white) in the edge and minimum 0 (black)}
    \label{fig:database sample}
\end{figure}

The \textit{simple} database contains $1000$ 2-plane corners where again camera is separated by 3 length units, but the two planes share the same material and the material is chosen at random from the six materials in Table 1. Value of the angle is restricted to $\alpha \in [\frac{\pi}{6},\frac{2\pi}{3}]$ and other camera position angles have been sampled between as follows: $\theta = \frac{\pi-\alpha}{2}$, $\phi \in [\frac{\pi}{6},\frac{2\pi}{3}] $, $\gamma=0$, where these are standard Euler angles. These present a subset of corners that are frequently encountered in real life scenes.

\section{Learning the ToF Correction}

\begin{figure}
    \centering
    \includegraphics[width=7cm,height=7.5cm]{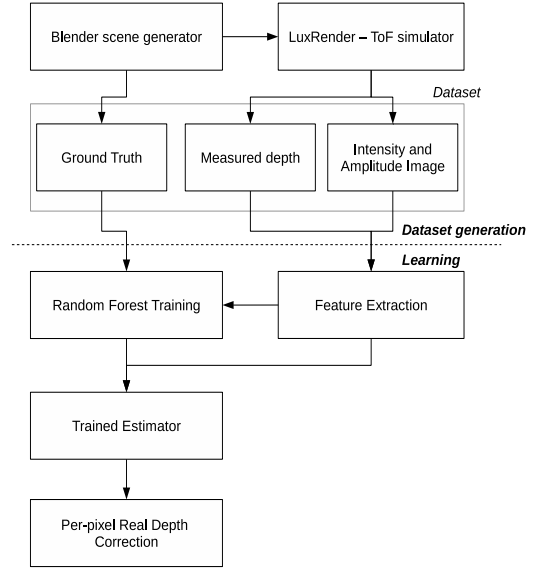}
    \caption{The pipeline for generating the corner database and learning the per-pixel correction. In learning stage, Random Forest regressor learned the real valued correction to depth image.}
    \label{fig:b}
\end{figure}

\subsection{Learning Pipeline}
As said in the introduction our aim is to predict a real valued correction to the pixel value based on the features of the pixel. We use a similar learning model as in \cite{Song2014} and \cite{Reynolds2011} of Random Forests. However in our case we use Regression Random Forests \cite{Criminisi2012} with simple linear separators. We used implementation of the algorithm from Scikit-learn \cite{scikit-learn}.

In order to asses whether our algorithm works we used relative per-pixel error as the main measure of error. We give definition of this quantity in the following equation. We store ground truth depth of a scene as a square matrix $D_{GT}$, measured depth from simulator as $D$, and lastly corrected depth as $D_C$. Hence, the relative per-pixel error at $i,j$ position in the matrix is:
\begin{equation}\label{eq:err}
\operatorname{RPE}_{ij}=\frac{|\operatorname{D_{GT}}-\operatorname{D}|}{|\operatorname{D_{GT}}|} \tag{Relative Per-Pixel Error}
\end{equation}
\subsection{Features}
We had three images at disposal from our LuxRender measurement. These include amplitude, intensity and depth images. We extracted features that previous studies \cite{Reynolds2011,Song2014} found important. In addition, we include some novel ones. These features could be separated to two categories local and spatial. In contrast to previous studies, we did not include the global features such as mean or median values of image depth as we believe these could create extremely biased model towards a specific scene i.e. a corner with 3.5 unit separation. Altogether we used 39 real-valued features. 

\textbf{Local Features} are the primary information that is used in many stages noise removal for individual pixels. These include \emph{Intensity}, 
\emph{Depth} and \emph{Amplitude} of the pixel. We also used radial distance of a pixel as it proved to be important in \cite{Song2014}. Additionally, we used a Gaussian measure of confidence denoted as $\mathcal{C}_{ij}$ as outlined in equation \eqref{eq:confidence} for each pixel. Confidence was used as a feature in similar fashion as in \cite{Reynolds2011}, however, in our analysis we chose not to include any distance bias. 
\begin{equation}\label{eq:confidence}
\resizebox{.8\hsize}{!}{$\mathcal{C}_{ij}=\exp\left({-\left(D_{ij}\cos\left(\frac{-\pi}{4}\right)+D_{ij}\sin\left(\frac{-\pi}{4}\right)\right)^2}\right)$}
\end{equation}
\textbf{Spatial Features} encode information about the immediate neighbourhood of the pixel. We used filters on both Intensity and Depth image. These included Laplacian filters and Canny filters with kernels 3 $\times$ 3, 5 $\times$ 5 and 7 $\times$ 7. These filters are used for edge detection and could signalize where an edge is located. Additionally, we used Gabor filters with orientation of 0, 45, 90 and 135 degrees (these are 13 $\times$ 13 filters). Gabor filters help in distinguishing the orientation of an edge. Next, we used gradient information in various direction (x,y and both at once). We included magnitude and angle of gradient as separate features as they posses additional information to the directional information. Lastly, we used Local Binary Patter (LBP) to extract information about textures \cite{Ojala1994} from intensity and depth image.

\section{Results \& Discussion}
\begin{figure}
    \centering
    \includegraphics[width=7cm,height=6cm]{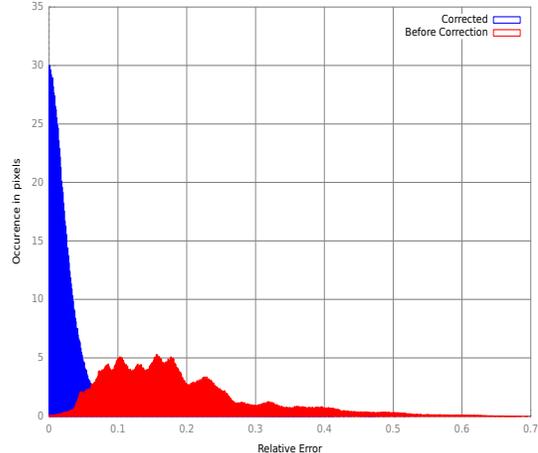}
    \caption{The relative per-pixel error distribution before and after the correction. We see that variance and mean of the distribution has been greatly improved. The statistics was done on $760,000$ pixels selected at random from the test set.}
    \label{fig:d}
\end{figure}
The learning part of the pipeline in Figure \ref{fig:b} was done via Regression Random Forest using python toolkit scikit-learn \cite{scikit-learn}. The Random Forest was generated with 150 classifiers with maximum depth 15 and minimal split $10,000$ on dataset that sampled $300$ images ($200 \times 200$) from \emph{simple} dataset. This accounts to $12, 000 ,000$ data points each with $38$ features. Given the size of dataset the use of word large-scale learning can be justified.

Testing of the dataset was done on $19$ randomly selected images that were not present in training dataset. This account to $ 760,000$ data points. The main measure of error was relative per-pixel error defined in the Equation \eqref{eq:err}. In Figure \ref{fig:d}, we can see the distribution of relative per-pixel error before and after the correction. We observe that the error in the sample has been great reduced. Statistical properties of the distribution have been improved with mean relative per-pixel error improving from $0.19$ to $0.031$. Additionally, variance of the distribution has been reduced by one order of magnitude from $0.012$ to $0.0015$. This hints us that we are not only subtracting mean of the error from the depth image. We selectively identify different types of pixels and scenes to achieve significantly better corrections. 

\begin{figure}
    \centering
    \includegraphics[width=7cm,height=5.5cm]{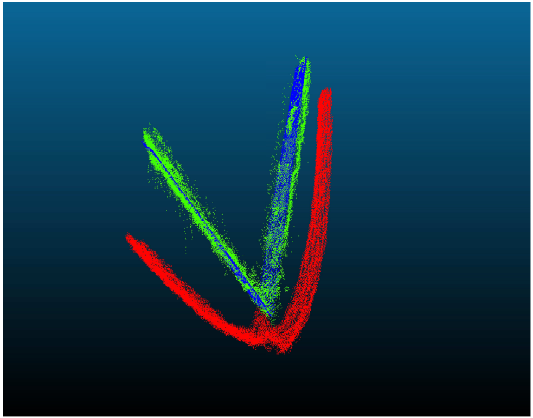}
    \caption{A point cloud of a corner scene. In  red one can see the measured depth with strongly pronounced multipath effect compared to blue ground truth image. In green we can see the corrected depth. There are some imperfections to the correction however the general trend is captured very well.}
    \label{fig:c}
\end{figure}

\begin{figure}
    \centering
    \includegraphics[width=7cm,height=6cm]{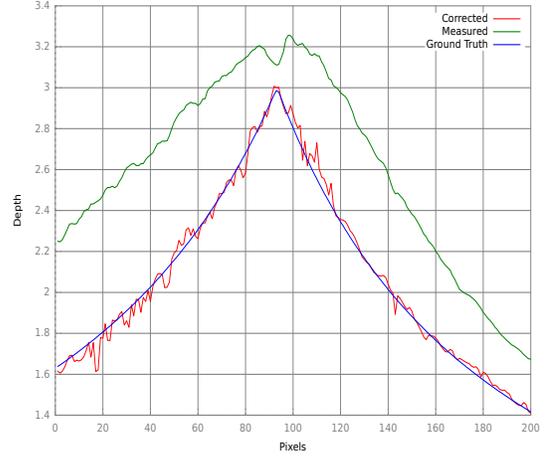}
    \caption{Cross-section of the same scene as in Figure \ref{fig:c}. We see that the general trend has been well captured.}
    \label{fig:e}
\end{figure}

When looking at Figures \ref{fig:c} and \ref{fig:e} we see that the general trend of the ground truth has been correctly identified by the learning model. We see that there are some errors to the correction, but at this point it could be argued that they are relatively small (mean being approx 3\%) and centred around ground truth depth of an image. As the bias from multipath has been already removed by our algorithm, these errors could be removed using variational denoising very easily.

\begin{figure}
    \centering
    \includegraphics[width=7cm,height=6cm]{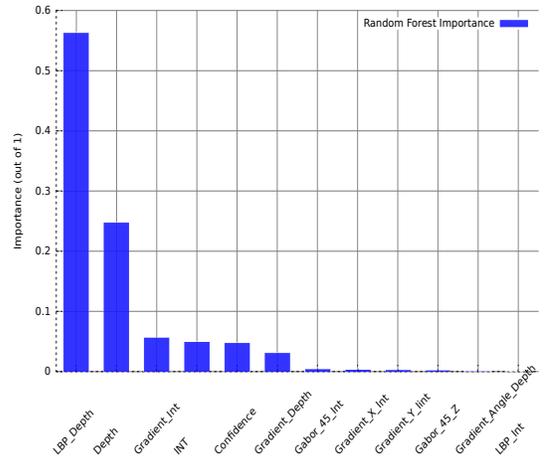}
    \caption{The first 12 (out of 39) most important features from the Random Forest learning.}
    \label{fig:f}
\end{figure}

Lastly, in Figure \ref{fig:f} we see that out of 38 features only 12 seem to be significantly important (above $0.001$) and have a non-negligible feature importance when calculated from occurrence in the Random Forest. Our feature importance is only with partial agreement in \cite{Reynolds2011} and \cite{Song2014}. We were suspecting that Depth, Intensity and Gabor features will be important due to results from aforementioned studies. Also, we suspected that LBP will be an important feature as it could hint something about textures of the surfaces. It is generally understood that the multipath effect is pronounced on specular surfaces the most, such as metallic surface. However, what seems to be eluding to us is why LBP of depth image was more important that LBP of intensity image. One would intuitively expect that intensity image should capture more material information.
\section{Conclusion}
In this work we presented a novel and easy way to mitigate multipath effect from Time-of-Flight cameras with experimental results on a newly generated dataset. We showed a great improvement in relative per-pixel error statistics with improving variance of relative error by an order of magnitude showing that a learning estimator has captured a difficult nature of the learning objective. We believe that our algorithm managed to eliminate the multipath bias on our dataset. Future work should focus on more challenging scenes and implementing real time analysis.
\bibliographystyle{plain}
\bibliography{Report_0_1}

\end{document}